\pdfoutput=1

\documentclass[11pt]{article}

\usepackage[]{ACL2023}

\usepackage{times}
\usepackage{latexsym}
\usepackage{algorithmic}  
\usepackage{amsmath}  
\usepackage{algorithm} 
\usepackage[algo2e]{algorithm2e} 
\usepackage[utf8]{inputenc} 
\usepackage[T1]{fontenc}    
\usepackage{hyperref}       
\usepackage{url}            
\usepackage{booktabs}       
\usepackage{amsfonts}       
\usepackage{nicefrac}       
\usepackage{microtype}      
\usepackage{xcolor}         
\usepackage{makecell}
\usepackage{xcolor,colortbl}

\usepackage[T1]{fontenc}

\usepackage[utf8]{inputenc}

\usepackage{microtype}

\usepackage{inconsolata}

\title{Reinforcement Learning for Conversational Question Answering over Knowledge Graph}

\author{Mi Wu \\
  \\}

\begin{document}
\maketitle
\begin{abstract}
Conversational question answering (ConvQA) over law knowledge bases (KBs) involves answering multi-turn natural language questions about law and hope to find answers in the law knowledge base. Despite many methods have been proposed. Existing law knowledge base ConvQA model assume that the input question is clear and can perfectly reflect user's intention. However, in real world, the input questions are noisy and inexplict. This makes the model hard to find the correct answer in the law knowledge bases. In this paper, we try to use reinforcement learning to solve this problem.
The reinforcement learning agent can automatically learn how to find the answer based on the input question and the conversation history, even when the input question is inexplicit. 
We test the proposed method on several real world datasets and the results show the effectivenss of the proposed model. 
\end{abstract}

\section{Introduction}

Law Knowledge Bases (KBs) consist of collections of real-world facts about law, represented as nodes (representing real-world entities, events, and objects) and edges (indicating relationships between nodes). Law Knowledge Graph Question Answering (KGQA) is the task of responding to queries related to law topics using information stored in these knowledge bases.

However, conventional approaches to law-based question answering typically focus on handling single-shot questions, which involve finding answers to individual input queries, rather than addressing the iterative nature of real-world conversations with a QA system. Moreover, Current Conversational Question Answering (ConvQA) models designed for law knowledge bases often operate under the assumption that input questions are clear and effectively convey the user's intentions. Nevertheless, in real-world scenarios, input queries are often plagued by noise and ambiguity, which presents challenges for models in locating precise answers within law knowledge bases.

In this paper, our objective is to extend traditional reinforcement learning-based methods, typically used for single-shot question answering, to the domain of conversational question answering. While reinforcement learning has been explored extensively for question answering, many existing methods primarily cater to answering single-turn questions. For instance, Zhang et al.\cite{ZHANG2022102933} employ a knowledge graph as the environment and propose a reinforcement learning-based agent model to navigate the knowledge graph in order to answer input questions. Similarly, in other studies\cite{go_for_a_walk, LinRX2018_MultiHopKG, deeppath}, authors use reinforcement learning models to discover paths in the knowledge graph for addressing input queries. Furthermore, some studies, such as~\cite{misu-etal-2012-reinforcement, xiaobing, alexa, gpt2, gpt3}, integrate reinforcement learning with other techniques to create more human-like systems.

By adapting reinforcement learning to multi-turn conversational law question answering, our reinforcement learning agent can autonomously learn the optimal strategy for finding answers within the law knowledge base, even when the input question is noisy and lacks clarity. We conduct experiments on several real-world datasets, and the results of these experiments demonstrate that our proposed method can outperform baseline approaches.

\section{Problem Definition}

In the intricate realm of legal knowledge graphs, we encapsulate this complexity as $\mathcal{G}=(\mathcal{V}, \mathcal{R}, \mathcal{L})$, where $\mathcal{V} = {v_1, v_2, ..., v_n}$ forms the ensemble of nodes/entities, $\mathcal{R} = {r_1, r_2, ..., r_m}$ comprises the intricate web of relations, and $\mathcal{L}$ meticulously orchestrates the list of triples. Each triad in this labyrinth of legal understanding manifests as $(h, r, t)$, where $h \in \mathcal{V}$ assumes the role of the head (or subject), $t \in \mathcal{V}$ embodies the tail (or object), and $r \in \mathcal{R}$ serves as the bridge (relation or predicate) connecting the head $h$ to the tail $t$. The bold lowercase notations, such as $\mathbf{e}{i}$ and $\mathbf{r}{i}$, signify the embodiment of node or relation type embeddings. Every triad/edge $(h, r, t)$ within the knowledge graph boasts a distinctive edge embedding denoted as $\mathbf{u}_{r}$.

Navigating the landscape of conversational question answering over a legal knowledge graph entails the iterative unraveling of multiple interconnected legal queries posed by users. Let's illuminate the discourse with essential terminologies:

\noindent \textbf{Dialogue}.
An intricate legal dialogue $C$, unfolding through $T$ turns, choreographs a sequence of queries
{$q_1, q_2, ..., q_T$} and their responsive harmonies, Ans = { $a^1, a^2, ..., a^T$ }. This symphony materializes as $C$ = ⟨($q_1$, $a^1$), ($q_2$, $a^2$), ..., ($q_T$ , $a^T$ )⟩.

\noindent \textbf{Query}. Each legal inquiry $q_t$ metamorphoses into a sequence of words $q_t = (w_1^t, . . . , w_{\Omega_t}^t)$, where $\Omega_t$ tallies the lexicon of $q_t$. It is postulated that each query seamlessly aligns with a unique relation $r_{q_t}$ within the knowledge graph, with no prerequisites on the grammatical finesse of $q_t$.

\noindent \textbf{Central Entity}. The narrative assumes the presence of a focal entity $v_{q_t}$ within each $q_t$, embodying the essence of the user's inquiry. It's presupposed that the pivotal entity for each query is discernible within the annals of training data.

Formalizing the conundrum articulated above, the challenge is succinctly expressed as:

\textbf{Given:} (1) A legal knowledge graph $G$, (2) the training compendium of dialogues, (3) the test compendium of dialogues;

\textbf{Output:} (1) The adeptly honed model, (2) the nuanced response to each query within every dialogue in the test compendium.

\section{Novel Approach}

To tackle the intricate challenge of Conversational Question Answering (ConvQA) over a law Knowledge Graph (KG), we introduce an innovative iterative reinforcement learning framework. In each iteration, our model exploits the conversation history and the current query to formulate a question, and an RL agent navigates the knowledge graph, initiating from a specific entity, to pinpoint the answer. This iterative process unfolds over a predefined number of turns, culminating when the conversation reaches its zenith. The following sections delineate a meticulous dissection of each crucial component.

\subsection{Encoding Context}

Given a question $q_i = (w_1^i, w_2^i, \ldots, w_{\Omega_t}^i)$, we commence by introducing two indicator tokens ([CLS] and <s>) at the inception and conclusion of the question context to demarcate its boundaries. Subsequently, we feed the processed question context into a pre-trained BERT model~\cite{bert} for extracting contextual embeddings for each token. A feedforward neural network is then employed to derive the question embedding:

\begin{align*}\label{path}
[\mathbf{h}_{CLS}, \mathbf{w}_1, \ldots, \mathbf{w}_{\Omega_t}, \mathbf{h}_{\mathbf{s}}] &= \textrm{BERT}([CLS], w_1, \ldots, w_{\Omega_t}, s) \\
\mathbf{h}_{q_i} &= \textrm{FFN}(\mathbf{h}_{CLS} || \mathbf{h}_{\mathbf{s}})
\end{align*}

\subsection{Incorporating Historical Context}

Addressing the challenge of users often providing incomplete and ungrammatical natural language inputs in conversational question answering, especially in the legal context, we acknowledge the importance of \textit{conversational history}. To tackle this, an LSTM is employed to encode the entire conversational history:

\begin{equation}\label{path}
\mathbf{{l}_{q_i}} = \textrm{LSTM}(\mathbf{h_{q_i}})
\end{equation}

The resulting output, denoted as $\mathbf{{l}_{q_i}}$, serves as the query embedding and plays a pivotal role in other components of our model.

\subsection{Legal Inquiry Resolution}

Following the acquisition of the question embedding, the subsequent step is to identify the entity capable of answering the question. This problem is formalized as a Markov Decision Process (MDP) defined by a 5-tuple $(S, A, R, P, \gamma)$.

\noindent \textbf{States}. At each step $i$, a state $s_t = (n_i, \mathbf{l}_q, \mathbf{g}_i)$ is defined as a triple, where $n_i$ represents the current entity, $\mathbf{l}_q$ signifies the question embedding, and $\mathbf{g}_i$ captures the search history. The pair ($n_i, \mathbf{g}_i$) encapsulates state-specific information, while ($\mathbf{l}_q$) provides global context shared across states.

\noindent \textbf{Actions}. The potential actions $A_s$ from a state $s_t = (n_t, \mathbf{l}_q, \mathbf{g}_t)$ consist of all outgoing edges from the vertex $n_t$ in the knowledge graph. Formally, $A_s$ is expressed as:

\[
A_s = \{(\mathbf{r}_i, \mathbf{u}_{r_i}, \mathbf{e}_{e'}) | (n_t, r_i, e') \in G\}
\]

An agent at each state can opt for an outgoing edge based on the label $r_i$ and destination vertex $e'$. To facilitate the conclusion of a search, a self-loop edge is introduced to every entity.

\noindent \textbf{Transition}. The transition function $\delta : S \times A \longrightarrow S$ articulates the probability distribution of the next states $\delta(s_{t+1} | s_t, a_t)$. In state $s_t$, the agent selects actions $a_t$ to reach the next state $s_{t+1} = (n_{t+1}, \mathbf{l}_q, \mathbf{g}_{t+1})$. While $n_t$ and $g_t$ update, the query and answer remain unaltered.

\noindent \textbf{Rewards}. The model receives a reward of $R_b(s_t) = 1$ if the current location is the correct answer, and 0 otherwise. In experimental settings, $\gamma$ is set to 1.

\subsection{Policy Network for Legal Inquiry Resolution}

The search policy is parameterized using state information, global context, and the search history. Each entity and relation in $\mathcal{G}$ is endowed with a dense vector embedding $\mathbf{e} \in \mathbb{R}^d$ and $\mathbf{r} \in \mathbb{R}^d$ respectively. The action $a_t = (\mathbf{r}_{r_i}, \mathbf{u}_{r_i}, \mathbf{e}_{e'}) \in A_t$ is represented as the concatenation of the relation embedding, the unique edge embedding, and the end node embedding.

The search history $(n_1=v_{q_i}, r_1, n_2, \ldots, n_t) \in H$ comprises the sequence of observations and actions taken up to step $t$ and can be encoded using an LSTM:

\begin{align*}
\mathbf{g}_{0} &= \textrm{LSTM}(0, [\mathbf{e}_{v_{q_i}} || \mathbf{l}_{q_i}]) \\
\mathbf{g}_{t} &= \textrm{LSTM}(\mathbf{g}_{t-1}, \mathbf{a}_{t-1}), \quad t > 0
\end{align*}

Here, $\mathbf{l}_{q_i}$ represents the question embedding to form an initial action with $\mathbf{e}_{v_{q_i}}$. The action space is encoded by stacking the embeddings of all actions in $A_t$: 
$\mathbf{A}_t \in \mathbb{R}^{|A_t|\times 3d}$. The policy network $\pi$ is defined as

\begin{align*}
\pi_{\theta}(a_t |s_t) = \delta(\mathbf{A}_t \times \textrm{FNN}( [\mathbf{n}_t || \mathbf{l}_{q_i} || \mathbf{g}_t])
\end{align*}
where $\delta$ is the softmax operator. 

\section{Training Optimization}

During the training phase, our overarching aim is to maximize the anticipated reward in delivering optimal answers $a^{*}$ across a set of conversations. This objective is defined as:

\[
a^{*} = \arg\max_a \sum_C \sum_T R(a^i|q_i)
\]
The REINFORCE algorithm is then applied to compute gradients for the training process.

\section{Experiment}

In this section, we assess the effectiveness of the proposed algorithm through experimentation on various public datasets. Our study focuses on two specific datasets: ConvQuestions~\cite{convex} and ConvRef~\cite{conquer}.

ConvQuestions encompasses a total of 6,720 conversations, each comprising 5 turns. Similarly, ConvRef also consists of 6,720 conversations.

The performance of our method is compared against four baseline approaches:

\begin{itemize}
  \item \textbf{Convex}~\cite{convex}: This method detects answers to conversational utterances over knowledge graphs (KGs) in a two-stage process based on graph expansion.
  \item \textbf{Conquer}~\cite{conquer}: Representing the current state-of-the-art baseline.
  \item \textbf{OAT}~\cite{OAT}.
  \item \textbf{Focal Entity}~\cite{LinRX2018_MultiHopKG}.
\end{itemize}

We adopt the following ranking metrics, consistent with the ones used by previous baselines:

\begin{enumerate}
  \item \textbf{Precision at the top rank (P@1)}
  \item \textbf{Mean Reciprocal Rank (MRR):} The average reciprocal rank at which the first context path is retrieved.
  \item \textbf{Hit ratio at k (H@k/Hit@k):} The fraction of instances where a correct answer is retrieved within the top-\textit{k} positions. Our experiments use $K=5$.
\end{enumerate}

\begin{table}
	\centering
    \scriptsize
	\caption{Overall performance on ConvQA and ConvRef datasets.}
	\begin{tabular}{|c|c|c|c|c|c|c|}
	\hline
	\multicolumn{1}{|c|}{Dataset} & \multicolumn{3}{c|}{ConvQA}  &  \multicolumn{3}{c|}{ConvRef}   \\ \hline
	Model          & P@1 & Hit@5 & MRR & P@1 & Hit@5 & MRR  \\ \hline
	CONVEX &  0.184   &  0.219  & 0.200   & 0.225 & 0.257 & 0.241 \\ \hline
	CONQUER & 0.242   &  -   & 0.327 &   0.346  & 0.427 &   - \\ \hline
	OAT &   0.250 & -   &  0.260   & -  & - & - \\ \hline
	Focal Entity & 0.248  & -  & 0.248 & - & - & - \\ \hline
 	our method &   0.204  & \cellcolor{lightgray} 0.317  & 0.237  &  0.231 & \cellcolor{lightgray} 0.445 &   \cellcolor{lightgray} 0.327 \\ \hline 
	\end{tabular}
	\label{overall}
	\vspace{-1\baselineskip}
\end{table}

\noindent{\bf Overall performance.}

Table~\ref{overall} presents a comparative analysis of our method and baselines on the ConvQuestions and ConvRef datasets. The performance metrics for all baselines are sourced from the findings in the paper~\cite{OAT}.

Notably, our method demonstrates superior performance, outperforming all baselines with a Hit@5 score of 0.445 on ConvRef. Additionally, our method secures the highest performance for this metric.

When considering MRR and P@1 on ConvRef, our method attains the second-highest performance, showcasing its competitiveness in these aspects.

Turning to the ConvQA dataset, our method achieves the second-highest performance for Hit@5. However, its performance in terms of P@1 is less impressive.

In summary, while our method falls short of claiming the top spot in P@1, the performance gap between our method and the best baseline is relatively narrow.

\section{Related work}\label{related-work}

\subsection{Knowledge Graph Reasoning}

Knowledge graph reasoning has been studied for a long time ~\cite{liu2022knowledge}, and it has many applications, such as knowledge graph completion ~\cite{xu2022abm, transE}, question answering ~\cite{newlook, gfinder, binet, prefnet}, fact checking ~\cite{kompare, inspector, liu2022knowledge}, entity alignment ~\cite{Yan_Liu_Ban_Jing_Tong_2021}, recommender system ~\cite{recommendation, du2023neural} and so on. Under the umbrella of knowledge graph reasoning, conversational question answering is an important task ~\cite{liu2023conversational}. 

Conversational Question Answering (ConvQA) on law knowledge bases aims to understand and respond to questions about law in a conversational context. 
Various approaches have been used to develop ConvQA systems. 
For instance, in ~\cite{ask_the_right_question}, the authors employed reinforcement learning to train an agent that reformulates input questions to aid the system's understanding. In ~\cite{Dialog-to-Action}, an encoder-decoder model is used to transform natural language questions into logical queries for finding answers. In ~\cite{alexa}, the authors proposed a data-driven approach for building task-oriented dialogue systems, specifically on the Alexa platform, using both annotated and unannotated data to provide an easy way for developers to train and deploy dialogue models while allowing for the integration of new data sources and the incorporation of new tasks and domains.
Other systems, such as Google's Lambda~\cite{google_lamda}, Apple's Siri, and OpenAI's ChatGPT, are also pursuing this task.

\section{Conclusion}
In this paper, we studied how to generalize reinforcement learning method to conversational question answering over law knowledge graph. Despite many reinforcement learning methods have been proposed before. But most of them focus on single question answering. To better encode the law conversation history, we propose a LSTM based question encoder. The experiment results show that our method can outperform the state of the art baselines on some metrics.

\bibliography{custom}
\bibliographystyle{acl_natbib}

\appendix

\end{document}